\documentclass[11pt,a4paper]{article}
\usepackage[hyperref]{eacl2021}
\usepackage{times}
\usepackage{latexsym}
\usepackage{epsfig}
\usepackage{graphicx}
\usepackage{xcolor}
\usepackage{url}

\usepackage{amsmath}
\usepackage{amssymb}

\usepackage{microtype}

\aclfinalcopy 
\def\aclpaperid{362} 

\title{UnibucKernel: Geolocating Swiss German Jodels Using Ensemble Learning}

\author{Mihaela G\u{a}man$^1$, Sebastian Cojocariu$^1$, Radu Tudor Ionescu$^{1,2,*}$\\
  $^1$Department of Computer Science, $^2$Romanian Young Academy\\
  University of Bucharest\\
  14 Academiei, Bucharest, Romania\\
  $^*$\texttt{raducu.ionescu@gmail.com} \\}

\date{}

\begin{document}
\maketitle
\begin{abstract}
In this work, we describe our approach addressing the Social Media Variety Geolocation task featured in the 2021 VarDial Evaluation Campaign. We focus on the second subtask, which is based on a data set formed of approximately 30 thousand Swiss German Jodels. The dialect identification task is about accurately predicting the latitude and longitude of test samples. We frame the task as a double regression problem, employing an XGBoost meta-learner with the combined power of a variety of machine learning approaches to predict both latitude and longitude. The models included in our ensemble range from simple regression techniques, such as Support Vector Regression, to deep neural models, such as a hybrid neural network and a neural transformer. To minimize the prediction error, we approach the problem from a few different perspectives and consider various types of features, from low-level character n-grams to high-level BERT embeddings. The XGBoost ensemble resulted from combining the power of the aforementioned methods achieves a median distance of $23.6$ km on the test data, which places us on the third place in the ranking, at a difference of $6.05$ km and $2.9$ km from the submissions on the first and second places, respectively. 
\end{abstract}

\section{Introduction}
\label{intro}

The Social Media Variety Geolocation (SMG) task was proposed, for the second year consecutively, in the 2021 edition of the VarDial Evaluation Campaign \cite{Chakravarthi-VarDial-2021}. This task is aimed at geolocation prediction based on short text messages exchanged by the users of social media platforms such as Twitter or Jodel. The location from where a short text was posted on a certain social media platform is expressed by two components: the latitude and the longitude. Naturally, the geolocation task is formulated as a double regression problem. Twitter and Jodel are the platforms used for data collection, and similar to the previous spin of SMG at VarDial 2020 \cite{Gaman-VarDial-2020b}, the task is divided into three subtasks, by language area, namely:

\begin{itemize}
    \item Standard German Jodels (DE-AT) - which targets conversations initiated in Germany and Austria in regional dialectal forms \cite{Hovy-EMNLP-2018}.
    \item Swiss German Jodels (CH) - containing a smaller number of Jodel conversations from the German speaking half of Switzerland \cite{Hovy-EMNLP-2018}.
    \item BCMS Tweets - from the area of Bosnia and Herzegovina, Croatia, Montenegro and Serbia where the macro-language is BCMS, with both similarities and a fair share of variation among the component languages \cite{Ljubesic-COLING-2016}.
\end{itemize}

The focus of our work falls only on the second subtask, SMG-CH, tackled via a variety of handcrafted and deep learning models. We propose a single ensemble model joining the power of several individual models through meta-learning based on Extreme Gradient Boosting (XGBoost) \cite{Chen-SIGKDD-2016}. We trained two independent ensemble models, each predicting one of the components that form the geographical coordinates (latitude and longitude). 

The first model plugged into our meta-learner is a Support Vector Regression (SVR) model \cite{Chang-NC-2002} based on string kernels. Previous usage in dialect identification has proved the efficiency of this technique in the task of interest \cite{Butnaru-VarDial-2018,Gaman-VarDial-2020,Ionescu-VarDial-2017,Ionescu-VarDial-2016}. 

The second model included in the ensemble is a hybrid convolutional neural network (CNN) \cite{Liang-2017} that combines, in the same architecture, character-level \cite{Zhang-NIPS-2015} and word-level representations. The ability of capturing morphological relationships at the character level and using them as features for CNNs is also known to give promising results in dialect identification \cite{Butnaru-ACL-2019,Tudoreanu-VarDial-2019}. Different from works using solely character-level CNNs for dialect identification \cite{Butnaru-ACL-2019,Tudoreanu-VarDial-2019}, we believe that the addition of words might bring the benefit of learning dialect-specific multi-word expressions that are hard to capture at the character level \cite{Dhingra-2016,Ling-2015}. 

Bidirectional Encoder Representations from Transformers (BERT) \cite{Devlin-NAACL-2019} is a top performing technique used in recent years for solving mainstream NLP problems. Thus, it seems fit to also include the outputs of a fine-tuned German version of BERT in our XGBoost meta-learner. 

We conducted experiments on the development set provided by the shared task organizers \cite{Hovy-EMNLP-2018} in order to decide which model to choose as our submission for the SMG-CH subtask. Our results indicate that the ensemble model attains the best performance. With median distances that are $5$-$6$ km higher, all the other models, tested individually on the development set, provide slightly worse predictions. 

The remainder of this paper is organized as follows. We present related work on dialect identification and geolocation of short texts in Section~\ref{sec_related}. Our approach is described in detail in Section~\ref{sec_method}. We present the experiments and empirical results in Section~\ref{sec_experiments}. Finally, our conclusions are drawn in Section~\ref{sec_conclusion}.

\section{Related Work}
\label{sec_related}

Our study of the related work starts with a brief overview of geotagging based on text. Then, we look at more specific methods studying geotagging in social media and we also investigate the amount of research done, from a computational linguistics perspective, in geotagging by dialect, with focus on last year's approaches for the same subtask, namely SMG-CH at VarDial 2020.

\noindent
\textbf{Text-based geotagging.} The works based on text to perform geotagging can be divided into three generic categories by the approaches taken in order to predict location. The first type of approach relies mainly on gazetteers as the source of the location mappings. This tool is adopted in a number of works \cite{Cheng-ACM-2010,Lieberman-ICDE-2010,Quercini-SIGSPATIAL-2010}, from ruled-based methods \cite{Bilhaut-NAACL-2003} to various machine learning techniques that rely on named entity recognition \cite{Ding-2000,Gelernter-2011,Qin-SIGSPATIAL-2010}. This category of methods brings the disadvantage of relying on specific mentions of locations in text, rather than inferring them in a less straightforward manner. These direct mentions of places are not a safe assumption, especially when it comes to social media platforms which represent the data source in some of these studies \cite{Cheng-ACM-2010}. The other two main categories of approaches for text-based geolocation rely on either supervised \cite{Kinsella-2011,Wing-HLT-2011} or unsupervised \cite{Ahmed-WWW-2013,Eisenstein-EMNLP-2010,Hong-WWW-2012} learning. The latter methods usually employ clustering techniques based on topic models.

\noindent
\textbf{Geolocation in social media.} A number of works \cite{Rout-ACM-2013} look at this task from a supervised learning perspective. However, in these studies, other details (e.g.~social ties) in the users profile are considered rather than their written content. Other works in this area are related to our current interest in studying language variation for the geolocation of social media posts \cite{Doyle-EACL-2014,Eisenstein-EMNLP-2010,Han-AIR-2014,Rahimi-Arxiv-2017,Roller-EMNLP-2012}. Among these, various machine learning techniques are employed in location prediction, ranging from probabilistic graphical models \cite{Eisenstein-EMNLP-2010} and adaptive grid search \cite{Roller-EMNLP-2012} to Bayesian methods \cite{Doyle-EACL-2014} and neural networks \cite{Rahimi-Arxiv-2017}. 

\noindent
\textbf{Dialect-based geolocation.} Many dialects are covered in the text-based geotagging research to date, including Dutch \cite{Wieling-PLoS-2011}, British \cite{Szmrecsanyi-IJHAC-2008}, American \cite{Eisenstein-EMNLP-2010,Huang-CEUS-2016} and African American Vernacular English \cite{Jones-AS-2015}. Out of all the languages that were subject to location detection by dialect, we are interested in German. In this direction, the study that is the most relevant to our work is that of \newcite{Hovy-EMNLP-2018} which targets the German language and its variations. Approximately 16.8 million online posts from the German-speaking area of Europe are employed in this study with the aim of learning document representations of cities. A small fraction of these posts are Jodels collected from the German speaking side of Switzerland, and these are also used in the SMG-CH subtask that we are addressing in this paper. The assumption here is that the methods should manage to capture enough regional variations in the written language, which can serve as the means to automatically distinguish the geographical region of social media posts. The verification performed in this direction in the original paper \cite{Hovy-EMNLP-2018} used clustering to determine larger regions covering a given dialect. However, given the shared task formulation, we take a different approach and use the provided data in a double regression setup, addressing the problem both from a shallow and a deep learning perspective.

As previously mentioned, this is the second consecutive year in which SMG-CH is featured at VarDial \cite{Chakravarthi-VarDial-2021}, with a similar format, although an updated data set. The participants of the 2020 SMG-CH shared task \cite{Gaman-VarDial-2020b} studied this task from a variety of angles. Some techniques are based on deep neural networks such as the popular BERT architecture \cite{Gaman-VarDial-2020,Scherrer-VarDial-2020}, bidirectional Long Short-Term Memory (LSTM) networks applied on FastText embeddings \cite{Mishra-VarDial-2020} or character-level CNNs \cite{Gaman-VarDial-2020}. Other techniques are based on shallow or handcrafted features such as a grid-based prediction using an n-gram language model \cite{Jauhiainen-VarDial-2020} or a clustering technique that shifts the problem into a discrete space, then uses an SVM for the classification of posts into regions. Our best submission \cite{Gaman-VarDial-2020} in last year's campaign was an ensemble based on XGBoost as meta-learner over the predictions of three different models: $\nu$-SVR with string kernels, a character-level CNN and an LSTM based on BERT embeddings. This year, we acknowledge that our deep learning models had sub-optimal results in our previous participation at SMG-CH. Consequently, for this year, we bring stronger neural networks into the XGBoost ensemble. Instead of using an LSTM with BERT embeddings, we fine-tune the cased version of German BERT and use it directly for double regression in a setup that is more suitable to the data set size, compared to our previous endeavour. Instead of a character-level CNN, we use a hybrid CNN which learns end-to-end representations for both words and characters. The aforementioned deep learning models are plugged into an XGBoost ensemble alongside a retrained version of the $\nu$-SVR model employed in our past participation \cite{Gaman-VarDial-2020}. Our current changes introduce a significant improvement in median distance compared to our previous results, on a similar (and perhaps more challenging) data set. 

\section{Method}
\label{sec_method}

\begin{figure*}[!t]
\begin{center}
\includegraphics[width=0.65\linewidth]{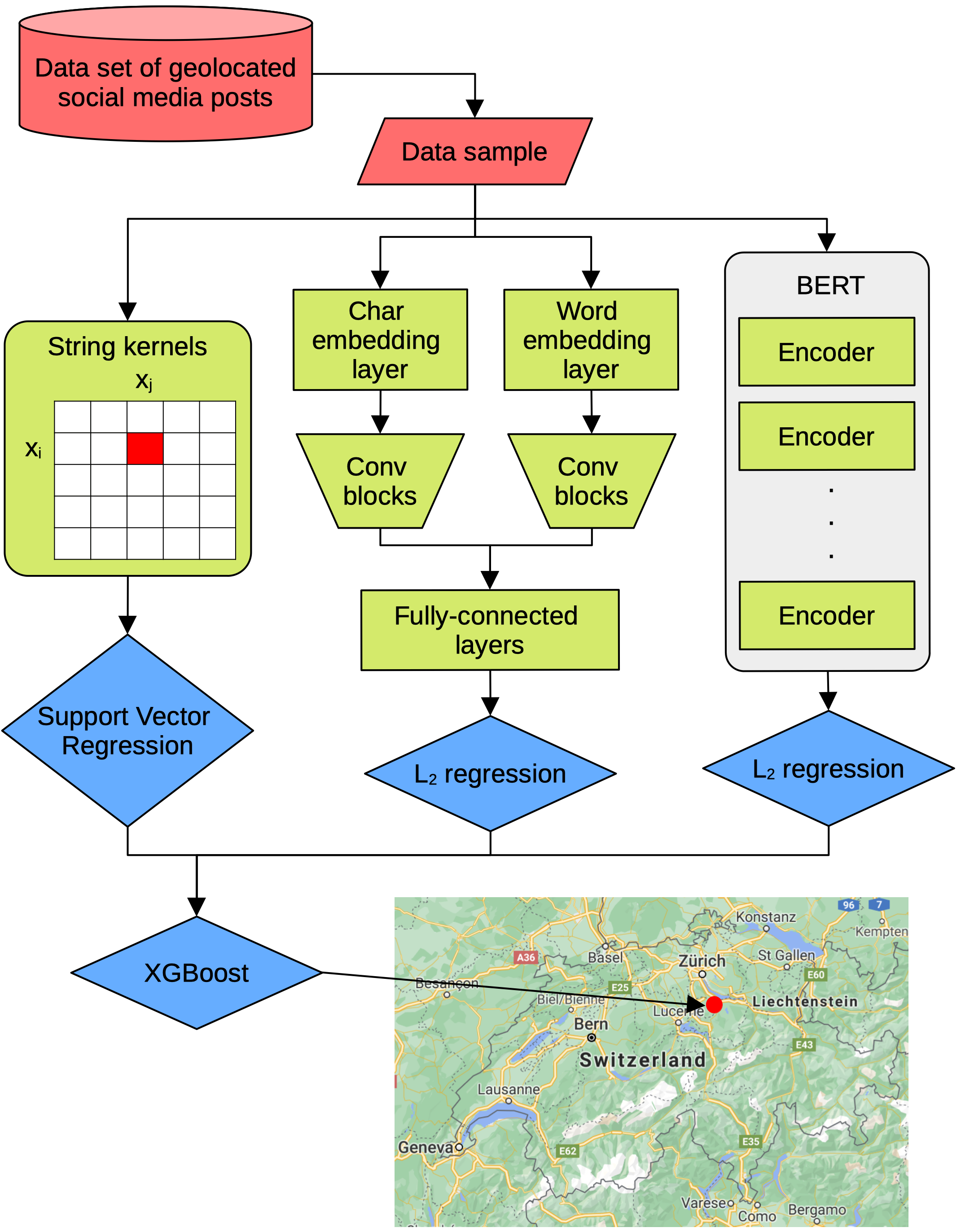}
\end{center}
\vspace*{-0.3cm}
\caption{Ensemble proposed by UnibucKernel for the SMG-CH shared task. Best viewed in color.}
\label{fig_pipeline}
\vspace*{-0.4cm}
\end{figure*}

The only system that our team submitted for the SMG-CH subtask is an ensemble model based on the XGBoost meta-learner, as illustrated in Figure~\ref{fig_pipeline}. In this section, we describe the three machine learning techniques that provide their predictions as input for the meta-learner, as well as the gradient boosting method that combines the independent models.

\subsection{Support Vector Regression with String Kernels}
\label{subsect:nusvr}

String Kernels \cite{Lodhi-NIPS-2001} provide a way of comparing two documents, based on the inner product generated by all substrings of length $n$, typically known as character n-grams. Being relatively simple to use and implement, this technique has many applications according to the literature \cite{Cozma-ACL-2018,Gimenez-EACL-2017,Masala-KES-2017,Ionescu-EMNLP-2018,Ionescu-EMNLP-2014,Ionescu-COLI-2016,Popescu-BEA8-2013}, with emphasis on dialect identification and the good results obtained for this task in previous VarDial evaluation campaigns \cite{Butnaru-VarDial-2018,Ionescu-VarDial-2017,Ionescu-VarDial-2016}.

Similar to our last year's submission for the SMG-CH subtask \cite{Gaman-VarDial-2020}, we employ the string kernels computed by the efficient algorithm introduced by \newcite{Popescu-KES-2017}. This gives us a dual representation of the data, through a kernel matrix where the cell on row $i$ and column $j$ represents the similarity between two text samples $x_i$ and $x_j$. In our experiments, we consider the presence bits string kernel \cite{Popescu-BEA8-2013} as the similarity function. For two strings $x_i$ and $x_j$ over a set of characters $S$, the presence bits string kernel is defined as follows:
\begin{equation}\label{eq_str_kernel_presence}
k^{0/1}(x_i, x_j)=\sum\limits_{g \in S^n} \mbox{\#}(x_i, g) \cdot \mbox{\#}(x_j, g),
\end{equation}
where $n$ is the length of n-grams and $\mbox{\#}(x,g)$ is a function that returns 1 when the number of occurrences of n-gram $g$ in $x$ is greater than 1, and 0 otherwise.

The resulting kernel matrix is plugged into a ${\nu}$-Support Vector Regression (${\nu}$-SVR) model. SVR \cite{Drucker-NIPS-1997} is a modified Support Vector Machines (SVM) \cite{Cortes-ML-1995} model that is repurposed for regression. Similar to SVM, SVR uses the notion of support vectors and margin in order to find an optimal estimator. However, instead of a separating hyperplane, SVR aims to find a hyperplane that estimates the data points (support vectors) within the margin with minimal error. In our experiments, we employ an equivalent SVR formulation known as $\nu$-SVR \cite{Chang-NC-2002}, where $\nu$ is the configurable proportion of support vectors to keep with respect to the number of samples in the data set. Using $\nu$-SVR, the optimal solution can converge to a sparse model, with only a few support vectors. This is especially useful in our case, as the data set provided for the SMG-CH subtask does not contain too many samples. Another reason to employ $\nu$-SVR in our regression task is that it was found to surpass other regression methods for other use cases, such as complex word identification \cite{Butnaru-BEA-2018}.

\subsection{Hybrid Convolutional Neural Network}
\label{subsect:charcnn}

Characters are the base units in building words that exist in the vocabulary of most languages. Among the advantages of working at the character level, we enumerate $(i)$ the neutrality with respect to language theory (independence of word boundaries, semantic structure or syntax) and $(ii)$ the robustness to spelling errors and words that are outside the vocabulary \cite{Ballesteros-EMNLP-2015}. These explain the growing interest in using characters as features in various language modeling setups \cite{Al-Rfou-AAAI-2019,Ballesteros-EMNLP-2015,Sutskever-ICML-2011,Wood-ICML-2009,Zhang-NIPS-2015}.

Word embeddings are vectorial word representations that associate similar vectors to semantically related words, allowing us to express semantic relations mathematically in the generated embedding space. From the initial works of \newcite{Bengio-JMLR-2003} and \newcite{Schutze-NIPS-1993} to the recent improvements in the quality of the embedding and the training time \cite{Collobert-ICML-2008,Mikolov-ICLRW-2013,Mikolov-NIPS-2013,Pennington-EMNLP-2014}, generating meaningful representations of words became a hot topic in the NLP research community. These improvements, and many others not mentioned here, have been extensively used in various NLP tasks \cite{Garg-PNAS-2018,Glorot-ICML-2011,Ionescu-NAACL-2019,Musto-ECIR-2016}.

Considering the sometimes orthogonal benefits of character and word embeddings, an intuitive idea has emerged, namely that of combining the character and word representations, which should complement each other in various aspects and provide better meaningful cues in the learning process of hybrid neural architectures \cite{Liang-2017}. Thus, throughout the experiments performed in this work, we choose to employ a hybrid convolutional neural network working at both the character level \cite{Zhang-NIPS-2015} and the word level \cite{Kim-EMNLP-2014}. The hybrid architecture concatenates two CNNs, out of which one is equipped with a character embedding layer and the other has an analogous word embeddings layer. The networks are able to automatically learn a 2D representation of text formed of either character or word embedding vectors, that are further processed by convolutional and fully-connected layers.

The last convolutional activation maps of our two CNNs sharing similar architectural choices are concatenated in what we call a hybrid network \cite{Liang-2017}, with the aim of accurately and simultaneously predicting the two location components required for the geolocation task. The first component of the hybrid network is a character-level CNN, which takes the first and last $250$ characters in the input and encodes each character with its position in the alphabet, then learns end-to-end embeddings for each character, as vectors of $128$ components. The second CNN used as part of the hybrid network operates at the word level and it receives as input each sample encoded, initially, as an array of $100$ indexes, corresponding to the position of each word in the vocabulary. As part of the pre-processing for the word-level CNN, we split the initial text into words, 
keeping the first $50$ words and the last $50$ words in the sample. In the end, we employ the German Snowball Stemmer \cite{Weissweiler-2018} to reduce each word to its stem, in an effort to reduce the vocabulary size by mapping variations of the same word to a single vocabulary entry. The word-level CNN is also equipped with an embedding layer, learning end-to-end word representations as vectors of length $128$. 

Each of the two CNNs has three convolutional (conv) layers placed after the initial embedding layer. The number of filters decreases from $1024$ for the first conv layer to $728$ for the second conv layer and to $512$ for the third conv layer. Each conv layer is equipped with Rectified Linear Units (ReLU) \cite{Nair-ICML-2010} as the activation function. The convolutional filter sizes differ across the two convolutional architectures. Hence, we use kernels of sizes $9$, $7$ and $7$ for the char CNN. In the same time, we choose $7$, $5$ and $3$ as appropriate filter sizes for the conv layers of the word CNN. In the char CNN, each conv layer is followed by a max-pooling layer with filters of size $3$. In the word CNN, we add max-pooling layers only after the first two conv layers. The pooling filter sizes are $3$ for the first pooling layer and $2$ for the second pooling layer. The activation maps resulting after the last conv blocks of the char and the word CNNs are concatenated and the hybrid network continues with four fully-connected (fc) layers with ReLU activations. The fc layers are formed of $512$, $256$, $128$ and $64$ individual neural units, respectively.

\subsection{Fine-Tuned German BERT}
\label{subsect:bert}

Transformers \cite{Vaswani-NIPS-2017} represent an important advance in NLP, with many benefits over the traditional sequential neural architectures. Based on an encoder-decoder architecture with attention, transformers proved to be better at modeling long-term dependencies in sequences, while being effectively trained as the sequential dependency of previous tokens is removed. Unlike other contemporary attempts at using transformers in language modeling \cite{Radford-Arxiv-2018}, BERT \cite{Devlin-NAACL-2019} builds deep language representations in a self-supervised fashion and incorporates context from both directions. The masked language modeling technique enables BERT to pre-train these deep bidirectional representations, that can be further fine-tuned and adapted for a variety of downstream tasks, without significant architectural updates. We also make use of this property in the current work, employing the Hugging Face \cite{Wolf-EMNLP-2020} version of the cased German BERT model\footnote{\url{https://huggingface.co/bert-base-german-cased}}. The model was initially trained on the latest German Wikipedia dump, the OpenLegalData dump and a collection of news articles, summing up to a total of $12$ GB of text files. We fine-tune this pre-trained German BERT model for the geolocation of Swiss German short texts, in a regression setup. The choice of hyperparameters is, in part, inspired by the winning system in the last year's SMG-CH subtask at VarDial \cite{Scherrer-VarDial-2020}.

\subsection{Ensemble Learning}
\label{subsect:xgboost}

Gradient tree boosting \cite{Friedman-2001} is based on training a tree ensemble model in an additive fashion. This technique has been successfully used in classification \cite{Li-UAI-2010} and ranking \cite{Burges-Learning-2010} problems, obtaining notable results in reputed competitions such as the Netflix Challenge \cite{Bennett-KDD-2007}. Furthermore, gradient tree boosting is the ensemble method of choice in some real-world pipelines running in production \cite{He-DMOA-2014}. XGBoost \cite{Chen-SIGKDD-2016} is a tree boosting model targeted at solving large-scale tasks with limited computational resources. This approach aims at parallelizing tree learning while also trying to handle various sparsity patterns. Overfitting is addressed through shrinkage and column subsampling. Shrinkage acts as a learning rate, reducing the influence of each individual tree. Column subsampling is borrowed from Random Forests \cite{Breiman-ML-2001}, bearing the advantage of speeding up the computations. In the experiments, we employ XGBoost as a meta-learner over the individual predictions of each of the models described above. We opted for XGBoost in detriment of average voting and a $\nu$-SVR meta-learner, both providing comparatively lower performance levels in a set of preliminary ensemble experiments.

\section{Experiments}
\label{sec_experiments}

\subsection{Data Set}
\label{sect:data}

The SMG-CH subtask \cite{Hovy-EMNLP-2018} offers, as support, a training set of 25,261 Jodel posts, provided in plain text format. Each textual input is associated with a pair of coordinates, i.e.~latitude and longitude, representing the position on Earth from where the text was posted. The development set is provided in an identical format and it is composed of 2,416 samples that we use to perform hyperparameter tuning and validate the results of our models. The test set has 2,438 samples without the corresponding coordinates, in order to avoid cheating or overfitting. Additionally, the proposed evaluation metric is the median distance between the predicted and the reference coordinates. A baseline median distance of $53.13$ km is also included in the specifications.

\subsection{Parameter Tuning}
\label{sect:tuning}

\noindent
\textbf{SVR based on string kernels.} 
We compute the presence bits string kernel using the efficient algorithm proposed by \newcite{Popescu-KES-2017}. In order to find the optimal range of n-grams, we experiment with multiple blended spectrum string kernels based on various n-gram ranges that include {n-grams} from 3 to 7 characters long. The best performance in terms of both mean absolute error (MAE) and mean squared error (MSE) was attained by a string kernel based on the blended spectrum of 3 to 5 character n-grams. These results are consistent with those reported by \citet{Ionescu-VarDial-2017} and \citet{Gaman-VarDial-2020}, suggesting that the 3-5 n-gram range is optimal for German dialect identification. The resulting kernel matrix is used as input for two $\nu$-SVR models, optimized for predicting the latitude and longitude (in degrees), respectively. For each of the two regressors, we tune the regularization penalty $C$, in a range of values from $10^{-4}$ to $10^{4}$. Similarly, for the proportion of support vectors $\nu$, we consider $10$ values uniformly covering the interval $(0, 1]$ with a step of $0.1$. To search for the optimal combination of hyperparameters, we apply grid search. Consistent with our previous study that inspired this choice of model and features \cite{Gaman-VarDial-2020}, for both regression models, the best hyperparameter combination is $C=10$ and $\nu=0.5$.

\begin{table*}[!th]
\begin{center}
\caption{The preliminary results of our team (UnibucKernel) obtained on the development set of the SMG-CH subtask.}
\label{tab_results_preliminary}
\vspace{-0.1cm}
\begin{tabular}{lcc}
\hline
Method                  & Median (km)       & Mean (km)     \\
\hline
Hybrid CNN              & 30.05             & 35.75         \\
$\nu$-SVR with string kernels              & 30.31             & 34.82         \\
BERT $\#1$ ($L_2$)             & 30.63             & 35.78         \\
BERT $\#2$ ($L_1$, truncated)             & 33.86             & 38.85         \\
BERT $\#3$ ($L_1$)             & 30.17             & 36.16         \\
\hline
XGBoost   ensemble      & 25.11             & 30.75         \\ 
\hline
\end{tabular}
\end{center}
\vspace{-0.2cm}
\end{table*}

\noindent
\textbf{Hybrid CNN.} 
As regularization for each conv block of the two CNNs, we introduce Gaussian noise and spatial dropout. We try many different magnitudes in the two regularization techniques, obtaining a few slightly different variations of the same model. These are compared against each other using grid search, selecting the best model to be included in our final ensemble. The rates used for the spatial dropout \cite{Tompson-IEEE-2015} are in the range $[0.001, 0.2]$, with a step of $0.001$, with the best dropout rate deemed to be $0.007$. For the Gaussian noise, we consider standard deviations in the range $[0.0001, 0.01]$, with the best results obtained with a standard deviation of $0.001$. After each of the four fully-connected layers, we employ plain dropout \cite{Srivastava-JMLR-2014} with a rate of $0.005$, deemed the best choice in the range $[0.001, 0.5]$.

In the optimization phase, common evaluation metrics such as the mean squared error (MSE) and the mean absolute error (MAE) are typically used to measure performance. In addition to these two metrics, we consider another error function as candidate for the loss to be minimized, namely the Huber loss. Although minimizing the Huber loss tends to give optimal values for the classical evaluation metrics, we finally use MSE as loss, given that it seems to converge to better results in terms of the median distance, which is the official evaluation metric in the SMG shared task. We optimize the hybrid CNN using the Adam optimization algorithm \cite{Kingma-ICLR-2014} with an initial learning rate of $10^{-3}$, chosen from the range $[10^{-5}, 10^{-2}]$, a weight decay of $10^{-7}$, selected in the initial range $[10^{-9}, 10^{-6}]$, and the learning rate decay of $0.999$. We train the hybrid architecture on mini-batches of $96$ samples for $1000$ epochs with early stopping. 
The network included in the final ensemble converged in $136$ epochs. 

\begin{table*}[!t]
\begin{center}
\caption{The final results obtained on the test set of the SMG-CH subtask by our ensemble model against the baseline and the top scoring methods.}
\label{tab_results_final}
\vspace{-0.1cm}
\begin{tabular}{lcccc}
\hline
Method                  & Rank & Median (km)       & Mean (km)     & AUC  \\
\hline
HeLju\_uc               & 1st & 17.55             & 25.84         & 74907.5   \\
HeLju\_c                & 2nd & 20.70             & 29.62         &  67237.5  \\
UnibucKernel            & 3rd & 23.60             & 29.75         &  63347.0    \\
\hline
baseline      & -        & 53.13             & 51.50         &  28637.0    \\

\hline
\end{tabular}
\end{center}
\vspace{-0.2cm}
\end{table*}

\noindent
\textbf{Fine-Tuned German BERT.} We fine-tune three BERT models in slightly different setups. The pre-trained base model used as starting point is the cased German BERT \cite{Wolf-EMNLP-2020}. We set the maximum sequence length to $128$, applying zero-padding to reach the fixed length for the samples that are shorter. The batch size, unanimously used, is $32$, with a training that goes on for a maximum of 50 epochs. We optimize the models using Adam, with an initial learning rate $\alpha=5\cdot10^{-5}$. The division by zero is prevented by the introduction of $\epsilon=10^{-8}$. Moving to the particularities of each of the three fine-tuned BERT models, we note a difference in the choice of the loss function, such that the first model employs the $L_2$ loss (MSE), while the other two models use the $L_1$ loss (MAE). Another variation is introduced by the text truncation choices, such that for the first and third models, we do not enforce truncation, while for the second model, all the samples are enforced to align to the provided maximum sequence length. These differences are reflected in the number epochs required for complete convergence: $20$, $18$ and $33$, respectively. We monitor the median distance for early stopping and observe that the best performance upon convergence is obtained by the third model, with a median distance of $30.17$ km on the validation set, followed by the results of the first ($30.63$ km) and second ($33.86$ km) models, respectively. In the ensemble, we include only the top scoring BERT model.

\noindent
\textbf{Extreme Gradient Boosting.} We employed XGBoost as a meta-learner, training it over the predictions of all the other models described in Section~\ref{sec_method}. We selected XGBoost as the submission candidate, as it provided the best results. The XGBoost regressor, deemed optimal by the hyperparameter tuning, has default values for most parameters\footnote{\url{https://xgboost.readthedocs.io/en/latest/}}, except for the number of estimators and the maximum tree depth. We set the number of estimators to $100$ for the latitude regressor and to $1000$ for the longitude regressor. Similarly, the maximum depth of the trees is $10$ for the latitude model and $20$ for the longitude one.

\subsection{Preliminary Results}
\label{sect:results}

We train three different models which rely on different learning methods and types of features to perform the required double regression. Thus, we have the hybrid CNN relying on both words and characters as features, the shallow $\nu$-SVR based on string kernels and three fine-tuned German BERT models looking at higher-level features and understanding dependencies in a bidirectional manner. 

Table \ref{tab_results_preliminary} shows the preliminary results obtained on the development set by each individual model as well as the results of the submitted XGBoost ensemble. The individual models provide quite similar results in terms of the median distance between the predicted and ground-truth locations. These results stay around a value of $30$ km for the median distance and $35$ km for the mean distance. Among the independent models, the hybrid CNN obtains slightly better results in terms of the median distance ($30.05$ km), whereas the second attempt at fine-tuning BERT gives the worst distances, namely $33.86$ km for the median distance and $38.85$ km for the mean distance. $\nu$-SVR surpasses all the other models, by a small margin, in terms of the mean distance ($34.82$ km). The results of the submitted XGBoost ensemble model stand proof that our intuition was correct, namely that all these individual models have the potential to complement each other if put together in an ensemble. Indeed, the submitted system clearly surpasses the best individual model by approximately $5$ km in terms of both the median and the mean distance metrics. 

\subsection{Final Results}

For our final submission, we trained all the individual models on both the provided training and development data sets. Then, we also retrained the submitted ensemble model, in a hope that this will give an even smaller median distance on the test set, compared to what we have obtained in the preliminary validation phase. 

Table~\ref{tab_results_final} shows an improvement in terms of the median distance on the test set compared to the one obtain on the development data. We cannot be sure that this improvement is solely due to the retraining that involves the development set. However, we conjecture that this endeavour played its part in the slightly better results. We outperform the baseline model by $29.53$ km in terms of the median distance and by $21.75$ km in terms of the mean distance, obtaining the third place in the competition. The constrained submission proposed by the organizers of the SMG-CH shared task surpasses our model by $2.9$ km in terms of the median distance and by $0.13$ km in terms of the mean distance. The unconstrained system on the first place, which was also proposed by the organizers of the SMG-CH shared task, distances itself by larger margins, with a difference of $6.05$ km for the median distance and a difference of $3.91$ km for the mean distance. 

\section{Conclusion}
\label{sec_conclusion}

In this paper, we proposed an ensemble learning model for the geolocation of Swiss German social media posts. The ensemble is based on an XGBoost meta-learner applied on top of three individual models: a hybrid CNN, an approach based on string kernels and a fine-tuned German BERT model. Given the final results obtained in the SMG-CH subtask, we conclude that predicting the location of Swiss German social media posts is a challenging task, the median distance being higher than 20 km. Using external data sources to build a language model seems to be a more promising path towards success, as shown by the final standings of the VarDial 2020 \cite{Gaman-VarDial-2020b} and 2021 \cite{Chakravarthi-VarDial-2021} SMG shared tasks.

In future work, we aim to study the applicability of our ensemble on other geolocation tasks, perhaps taking into consideration future VarDial challenges.

\section*{Acknowledgments}

The authors thank reviewers for their useful remarks.
This work was supported by a grant of the Romanian Ministry of Education and Research, CNCS - UEFISCDI, project number PN-III-P1-1.1-TE-2019-0235, within PNCDI III. This article has also benefited from the support of the Romanian Young Academy, which is funded by Stiftung Mercator and the Alexander von Humboldt Foundation for the period 2020-2022.

\bibliography{refs}
\bibliographystyle{acl_natbib}

\end{document}